\documentclass{spie}

\usepackage{graphicx}
\usepackage[cmex10]{amsmath}
\usepackage{amsfonts} 
\usepackage{amssymb}
\usepackage{stmaryrd}
\usepackage{subfigure}
\usepackage{algorithm,algorithmic}
\usepackage{multirow}
\usepackage{array}
\usepackage{url}
\usepackage[usenames,dvipsnames]{pstricks}

\newcommand{\be}{\begin{equation}}
\newcommand{\ee}{\end{equation}}

\title{Turbulence stabilization}
\author{Yu~Mao$^1$ and~J\'er\^ome~Gilles$^2$ 
\skiplinehalf
$^1$ Institute for Mathematics and Its Applications, University of Minnesota, 425 Lind Hall 207 Church Street SE, Minneapolis, MN 55455-0134, USA
\skiplinehalf
$^2$ Department of Mathematics, University of California Los Angeles, 520 Portola Plaza, Los Angeles, CA 90095-1555, USA
}

\authorinfo{Further author information: (Send correspondence to J\'er\^ome Gilles)\\
Yu Mao: maoxa004@ima.umn.edu\\
J\'er\^ome Gilles: jegilles@math.ucla.edu, Telephone: 1 310 794 7737, http://www.math.ucla.edu/$\sim$jegilles}

\begin{document}

\maketitle

\begin{abstract}
We recently developed a new approach to get a stabilized image from a sequence of frames acquired through atmospheric turbulence. The goal of this 
algorihtm is to remove the geometric distortions due by the atmosphere movements. This method is based on a variational formulation and is efficiently 
solved by the use of Bregman iterations and the operator splitting method. In this paper we propose to study the influence of the choice of the regularizing term 
in the model. Then we proposed to experiment some of the most used regularization constraints available in the litterature.
\end{abstract}


\section{Introduction}

These last few years show an increase of interest in the development of mitigation algorithms to deal with the atmospheric turbulence degradations.
Indeed, turbulences can affect images in two major ways: a blurring effect and random geometric distortions.\\
An interesting work about turbulence modelization for mitigation algorithms was made by Frakes \cite{Frakes:2001p7017,Gepshtein:2004p7030}. The authors 
modeled the turbulence phenomenon by using two operators:
\begin{equation}\label{eq:turbulence}
f_i(x)=D_i(H(u(x)))+\text{noise}
\end{equation}
where $u$ is the static original scene we want to retrieve, $f_i$ is the observed image at time $i$, $H$ is a blurring kernel, and $D_i$ is an operator 
which represents the geometric distortions caused by the turbulence at time $i$. Based on this model, the idea is to try to inverse the two operators $H$ 
and $D_i$. Inversing $H$ is a deconvolution problem which will not be addressed here (we propose an original deconvolution approach in \cite{Gilles:Fried}), in this paper we will focus on the stabilization problem to remove 
the geometric distortions.\\

In a recent work \cite{Mao:turb}, we proposed a new variational framework based on a combination of a deformation flow estimation and a nonlocal regularization 
term to retrieve a stabilized image from a set of acquired frames. Our results clearly outperform other existing methods like the PCA-based algorithm 
\cite{Li2007} or the Lucky-Region Fusion approach \cite{Aubailly2009}. If our algorihtm is very efficient and needs very few images (20 input frames seem
enough in all the cases), its main drawback is the computational time needed to apply the nonlocal regularization. In this paper, we propose to study 
the impact, in terms of image quality reconstruction, the choice of other regularizers.\\

The reminder of the paper is as follows. In section~\ref{sec:basicmodel} we recall the model developped in \cite{Mao:turb}. 
Section~\ref{sec:regularization} presents the different regularizer we choose to test instead of the nonlocal one. Corresponding experiments are presented in 
section~\ref{sec:results} and some conclusions are provided in section~\ref{sec:conclusion}.

\section{The stabilization Model}\label{sec:basicmodel}

In \cite{Mao:turb}, we proposed to use the following variational model (\ref{BasicOptimization}) to find a restored image from a set of input frames.
\be\label{BasicOptimization}
\min_{u,\phi_i} J(u)\quad\text{s.t.}\quad f_i=\Phi_i u+\text{noise},\quad \forall i
\ee

\begin{figure}
\begin{center}
\includegraphics[scale=0.6]{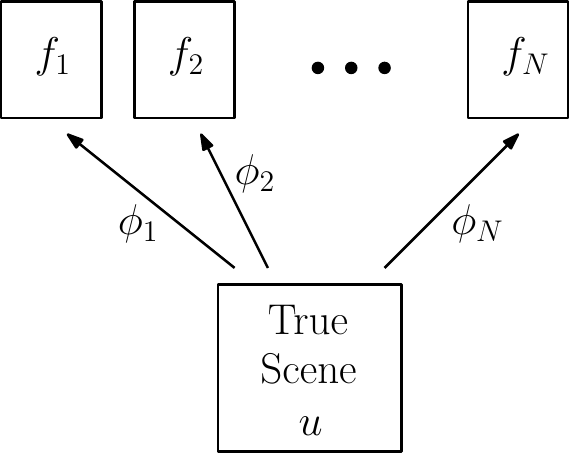}
\end{center}
\caption{The model of deformation used in the restoration process.}
\label{fig:deformmodel}
\end{figure}

Where the observed image sequence is denoted $\{f_i\}_{i=1,\ldots,N}$ and $u$ the true image that needs to be reconstructed.
Each $\phi_i$ corresponds to the geometric deformation on the $i$-th frame (note that the $\phi_i$ are the deformations between the true image and 
the observed frame $i$ and not the continuous movement flow from frame to frame, see Fig.~\ref{fig:deformmodel}). The term $J(u)$ is the regularizer 
applied to the reconstructed image, we will discuss its choice in the next section.

Such models can be efficiently solved by the alternative optimization method, i.e. optimizing over the different $u,\phi_i$ variables alternatively.

In our model \eqref{BasicOptimization}, if we have a good guess on $u$, then the optimal $\phi_i$ can be estimated via certain optical 
flow algorithms (e.g. the methods developed in \cite{Black:1996p5277,Bouguet:2000,Sun:2008p5795}). 
On the other hand, for fixed $\{\phi_i\}$ and by setting $\sum_i\|\Phi_i u-f_i\|_2^2$ be the data fidelity term, the model \eqref{BasicOptimization} 
can be rewritten as a constrained problem and efficiently solved via Bregman Iterations \cite{Osher:2005p632} (see \cite{Mao:turb} for more details 
about the model). 

The overall Algorithm \ref{OverallAlgorithm} is as follows:

\begin{algorithm}
\caption{The Alternative Optimization Algorithm}\label{OverallAlgorithm}
\begin{flushleft}
\begin{itemize}
\item[] Initialize: Start from some initial guess $u$. Let $\tilde f_i=f_i$.\\
\item[] \textbf{while} {$\sum_i\|\Phi_iu-f_i\|^2$ not small enough} \textbf{do}\\
\begin{itemize}
\item[] Estimate $\Phi_i$ which maps $u$ onto $f_i$ via optical flow.\\
\item[] \textbf{while} {$\sum_i\|\Phi_iu-\tilde f_i\|^2$ not converge} \textbf{do}\\
\begin{itemize}
\item[] $v\leftarrow u-\delta \sum_i\Phi_i^\top(\Phi_i u-\tilde f)$\\
\item[] $u \leftarrow \arg\min_u J(u)+\frac{\lambda}{2\delta}\|u-v\|^2$\\
\end{itemize}
\item[] \textbf{end while}\\
\item[] $\tilde f_i \leftarrow \tilde f_i+f_i-\Phi_i u$.\\
\end{itemize}
\item[] \textbf{end while}\\
\end{itemize}
\end{flushleft}
\end{algorithm}

\begin{figure}[!t]
\begin{center}
\includegraphics[width=0.2\textwidth]{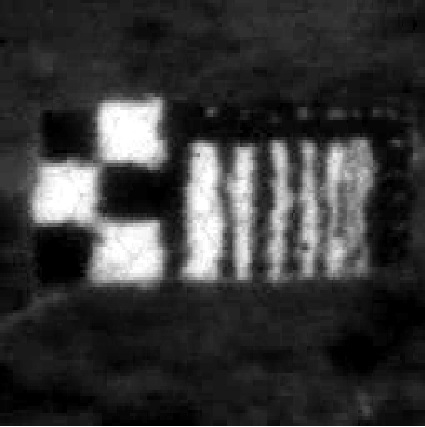}
\includegraphics[width=0.2\textwidth]{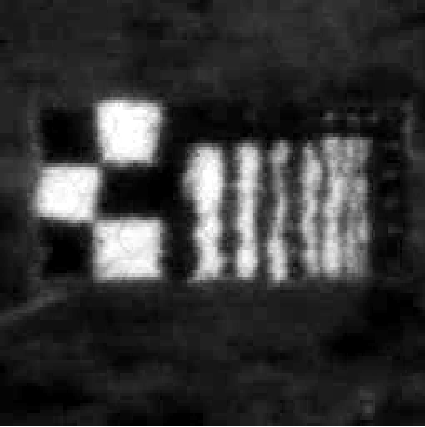}
\includegraphics[width=0.2\textwidth]{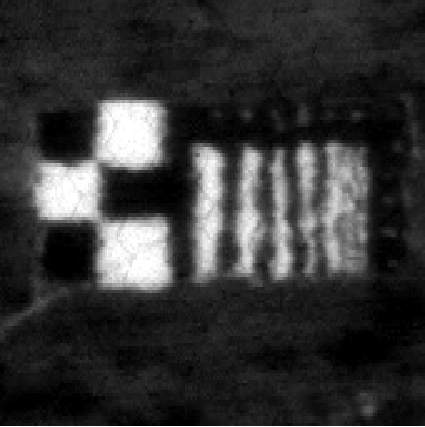}\quad
\includegraphics[width=0.2\textwidth]{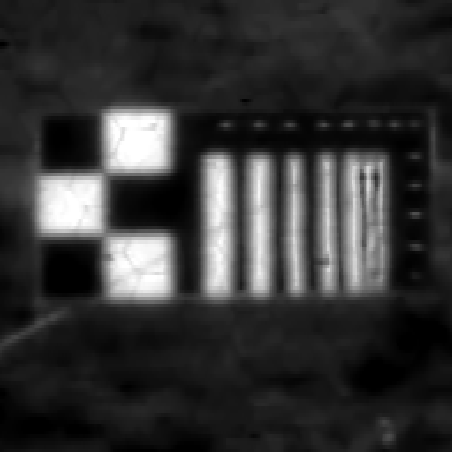}\\
\includegraphics[width=0.2\textwidth]{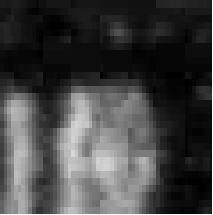}
\includegraphics[width=0.2\textwidth]{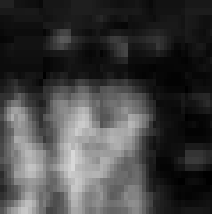}
\includegraphics[width=0.2\textwidth]{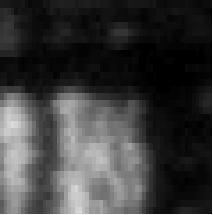}\quad
\includegraphics[width=0.2\textwidth]{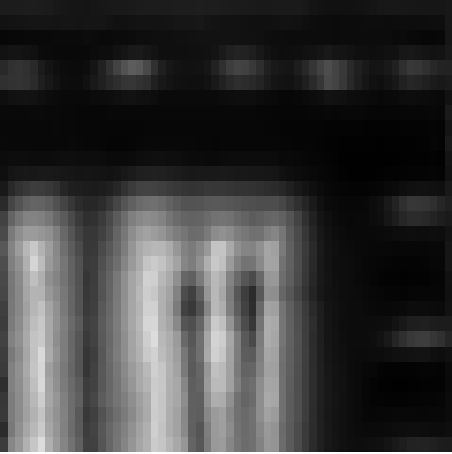}
\end{center}
\caption{The first three columns are example frames and the magnification of their top right part. The last column shows our reconstructed result.}
\label{example2}
\end{figure}

The initial value of $u$ is chosen as the temporal average of the frames; it is very blurry but gives a good initial guess of the rough shape of 
the object. The quantity of frames determines the reconstruction quality. On the other hand, the more frames used, the longer the computational 
time required is. In our numerical experiments we generally observe that we can always obtain satisfactory results with only 10-30 frames. We also note 
that the choice of the optical flow scheme is not crucial. Experiments made via the use of the Black-Anandan algorithm, which provides a precise 
flow estimation but slower, or the classic Lukas-Kanade algorithm, which is faster, didn't show notable differences. As we are interested in a fast 
algorithm, we finally retain the Lukas-Kanade scheme in the rest of the paper. Figure~\ref{example2} shows an example of the achieved result by this 
algorithm initially presented in \cite{Mao:turb}.

\section{Choice of the regularization}\label{sec:regularization}

As showed in the previous section, the algorithm contains a regularization step 
\begin{equation}\label{eq:regu}
u \leftarrow \arg\min_u J(u)+\frac{\lambda}{2\delta}\|u-v\|^2
\end{equation}

We will now discuss the choice of $J(u)$. In our original work \cite{Mao:turb}, we choose to use the nonlocal total variation (NLTV) as the regularizing 
term because it was successfully used in different restoration problems and is well adapted for real images. In order to clarify our discussion, we recall the definition of NLTV.

\begin{equation}
J(u)=J_{NLTV}(u)=\int_{\Omega}\sqrt{\int_{\Omega}\left(u(y)-u(x)\right)^2w(x,y)dy}dx
\end{equation}
where the weight $w(x,y)$ corresponds to a similarity measure between patches centered on pixel $x$ and $y$. As the similarity between the patches 
increases, so does their impact on the regularized image. It was proven that NLTV is well-adapted for images having textures or self-similarity. 
The main drawback of this regularizer is the time needed to compute the weight $w(x,y)$. In fact, the idea is to compare different patches all over the 
image which represents a huge number of combinations. While designing a new faster way to compute $w(x,y)$ is completely out of the scope of this paper, 
we propose to study the impact of choosing other regularizers. Nowadays, some regularizers are known to be efficients and considered as ``classic'' in the 
litterature. In addition to the NLTV, we choose three other constraint terms. The probably most known in image processing is the standard total variation 
(TV) proposed by Rudin et al. \cite{Rudin:1992p2320} which preserves sharp edges in the image; it is defined by (\ref{eq:rof}).

\begin{equation}\label{eq:rof}
J(u)=J_{TV}(u)=\int_{\Omega}|\nabla u|
\end{equation}

The other two choices are based on the idea that a regularized image will have a sparse representation in some ``dictionary''. This concept is directly related 
to the compressive sensing theory. Here we adopt two possible tight frames as our dictionaries: framelet \cite{Cai,Cai2009b} and curvelets 
\cite{Candes2003a,Candes2003b,Candes2005,Donoho1999a}. If we denote $\mathcal{D}$ and $\mathcal{C}$ the operators which respectively perform the framelet and curvelet expansions, the 
corresponding regularizers are defined by
\begin{equation}
J(u)=J_D(u)=\|\mathcal{D}u\|_1 \qquad \qquad \text{and} \qquad \qquad J(u)=J_C(u)=\|\mathcal{C}u\|_1
\end{equation}
Then depending on the choice of $J(u)$, the regularization step depicted by (\ref{eq:regu}) is equivalent to
\begin{itemize}
\item if $J(u)=J_{TV}(u)$, it is the famous Rudin-Osher-Fatemi model (ROF),
\item if $J(u)=J_{NLTV}(u)$, it is the nonlocal total variation presented at the beginning of this section,
\item if $J(u)=J_{D}(u)$ or $J(u)=J_{C}(u)$, it is an $L^1-L^2$ minimization problem widely studied in the litterature.
\end{itemize}

All of these minimization problems can be efficiently solved by the use of the Split Bregman Iterations. Corresponding Matlab routines are available in the 
Bregman Cookbook \cite{Gilles:BC} or in \cite{Zhang:2010p624} for NLTV.

\section{Experiments}\label{sec:results}

\begin{figure}[!t]
\begin{center}
\includegraphics[width=0.24\textwidth]{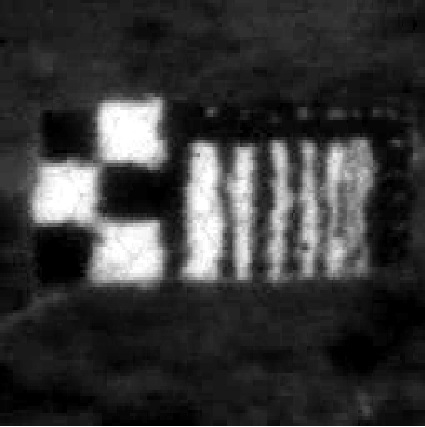}
\includegraphics[width=0.24\textwidth]{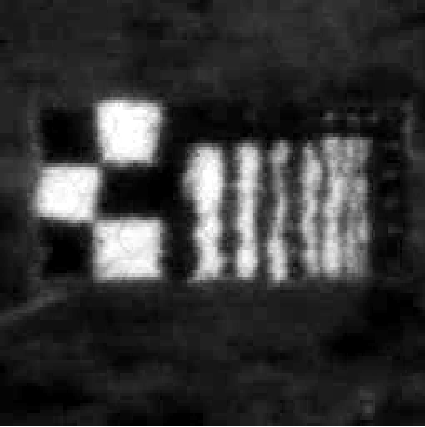}
\includegraphics[width=0.24\textwidth]{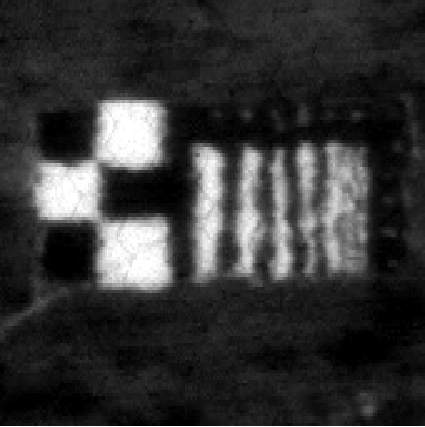}
\end{center}
\caption{Input examples of the test sequence 1.}
\label{in1}
\end{figure}

\begin{figure}[!t]
\begin{center}
\includegraphics[width=0.24\textwidth]{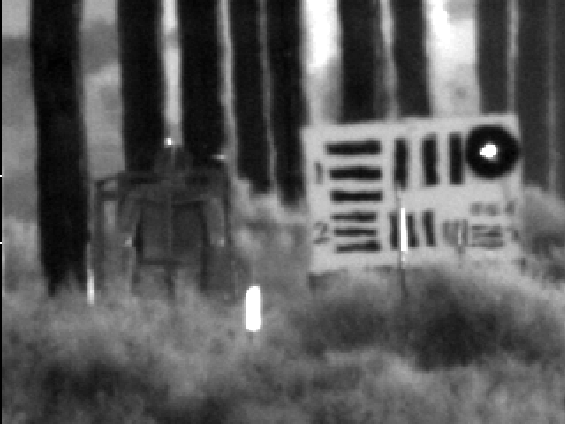}
\includegraphics[width=0.24\textwidth]{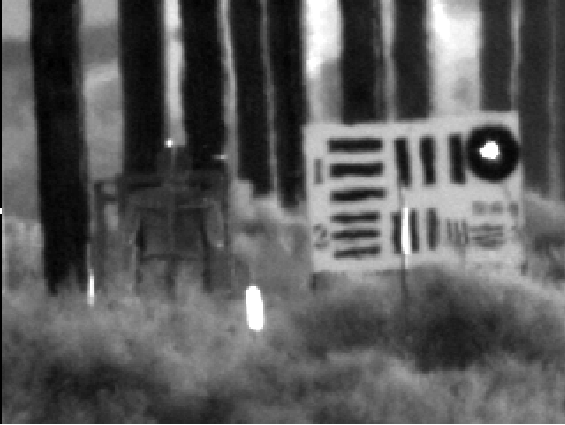}
\includegraphics[width=0.24\textwidth]{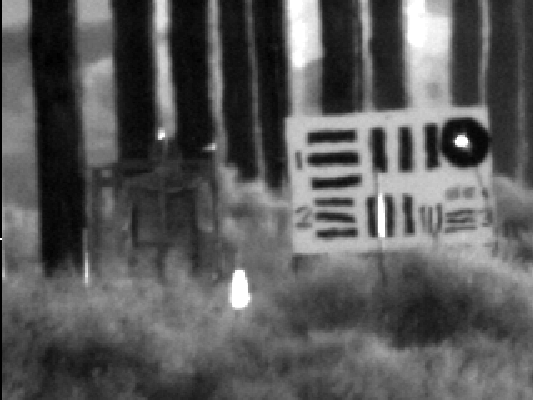}
\end{center}
\caption{Input examples of the test sequence 2.}
\label{in2}
\end{figure}

\begin{figure}[!t]
\begin{center}
\includegraphics[width=0.24\textwidth]{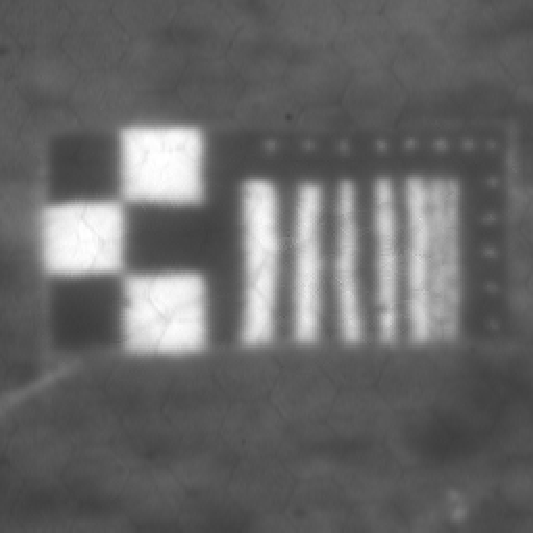}
\includegraphics[width=0.24\textwidth]{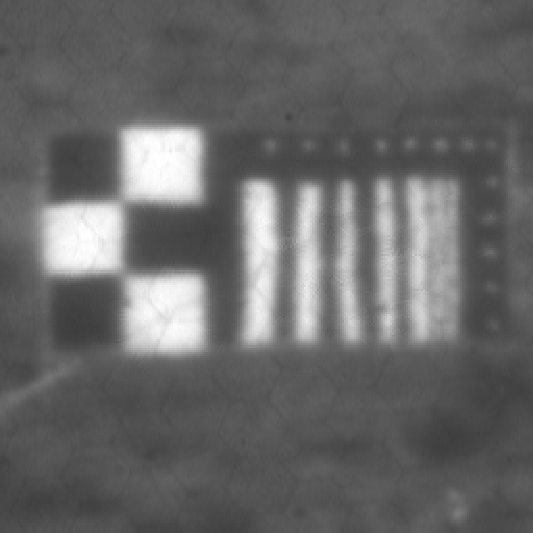}
\includegraphics[width=0.24\textwidth]{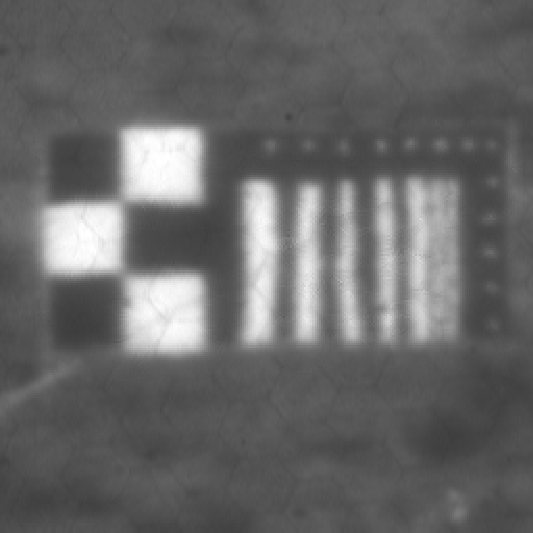}
\includegraphics[width=0.24\textwidth]{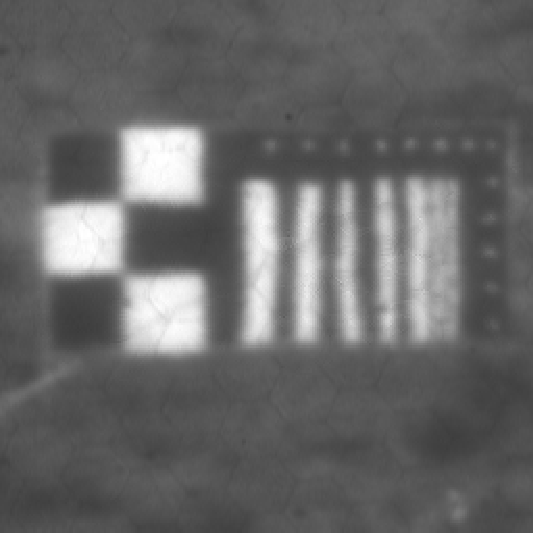}
\end{center}
\caption{Restored images obtained from the different regularizers (30 input frames). From left to right: NLTV, TV, Framelet, Curvelet.}
\label{example3}
\end{figure}

\begin{figure}[!t]
\begin{center}
\includegraphics[width=0.24\textwidth]{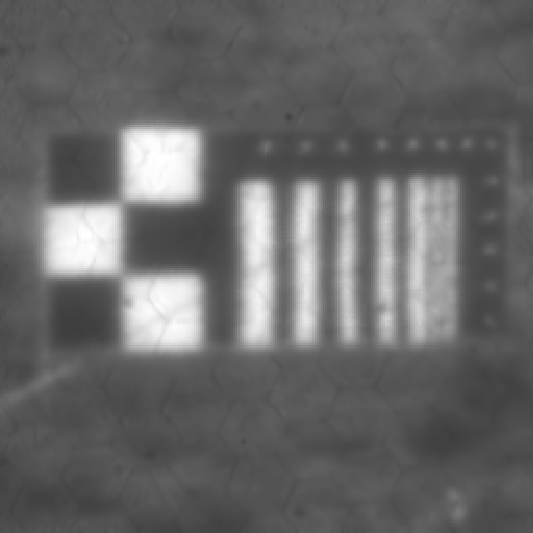}
\includegraphics[width=0.24\textwidth]{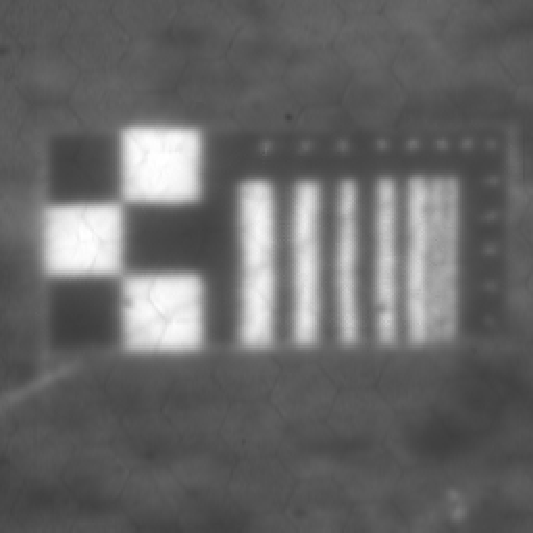}
\includegraphics[width=0.24\textwidth]{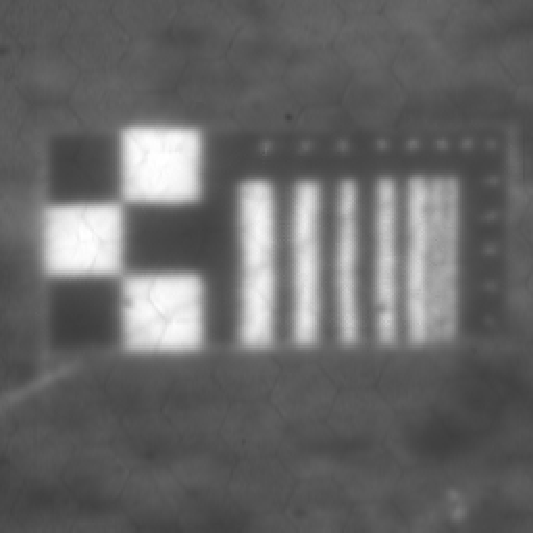}
\includegraphics[width=0.24\textwidth]{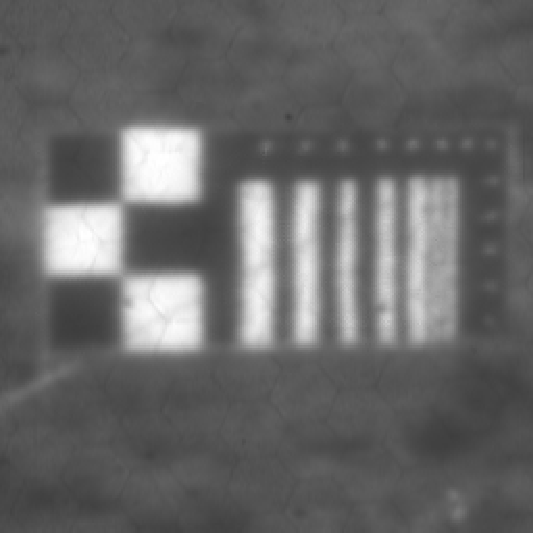}
\end{center}
\caption{Restored images obtained from the different regularizers (100 input frames). From left to right: NLTV, TV, Framelet, Curvelet.}
\label{example4}
\end{figure}

\begin{figure}[!t]
\begin{center}
\includegraphics[width=0.24\textwidth]{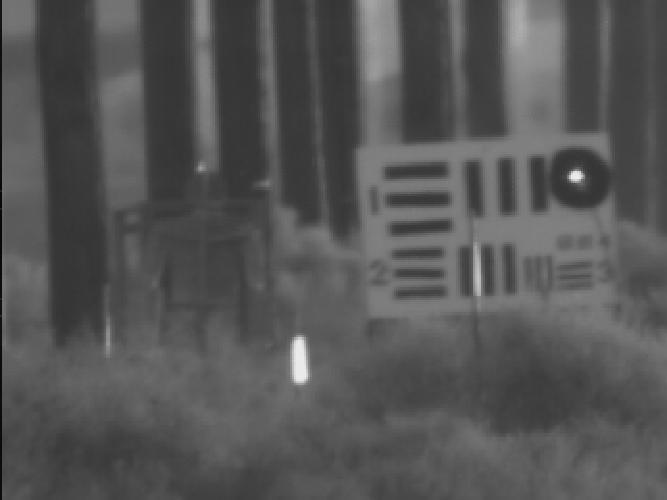}
\includegraphics[width=0.24\textwidth]{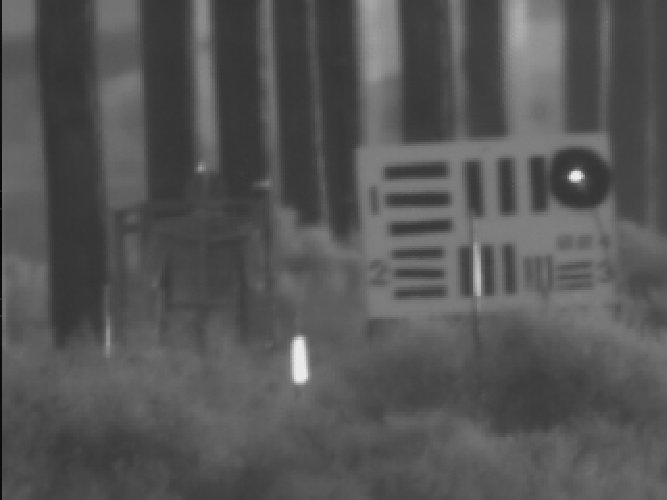}
\includegraphics[width=0.24\textwidth]{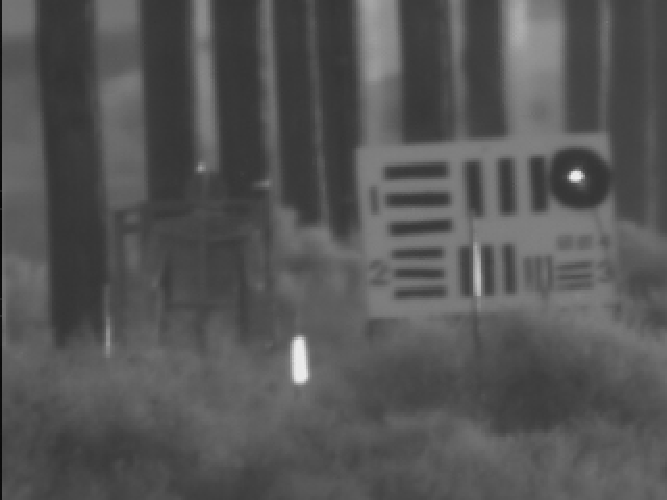}
\includegraphics[width=0.24\textwidth]{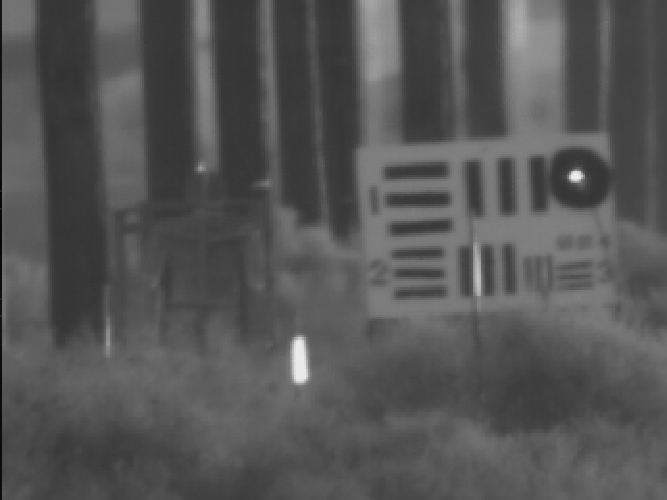}
\end{center}
\caption{Restored images obtained from the different regularizers (10 input frames). From left to right: NLTV, TV, Framelet, Curvelet.}
\label{example5}
\end{figure}

\begin{figure}[!t]
\begin{center}
\includegraphics[width=0.24\textwidth]{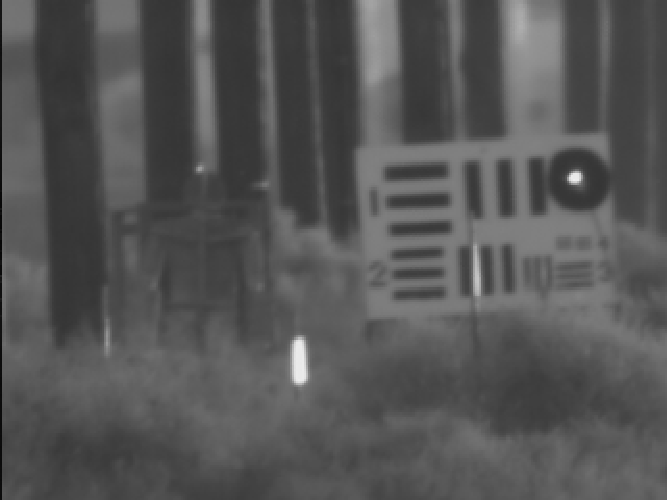}
\includegraphics[width=0.24\textwidth]{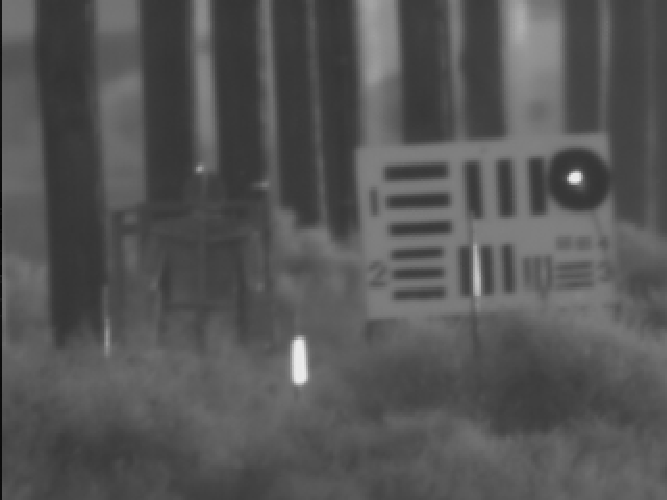}
\includegraphics[width=0.24\textwidth]{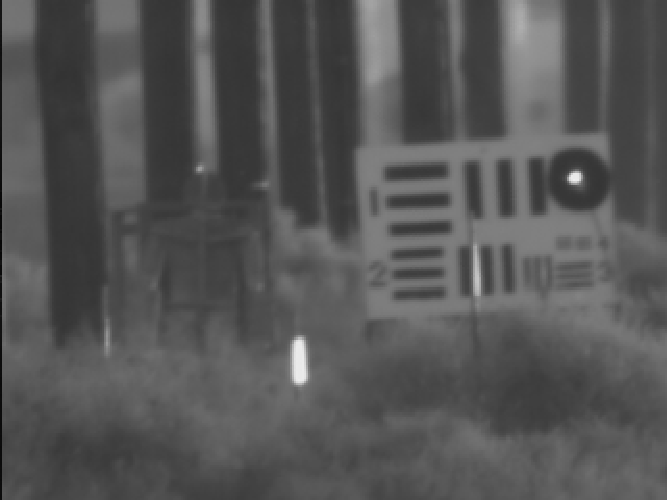}
\includegraphics[width=0.24\textwidth]{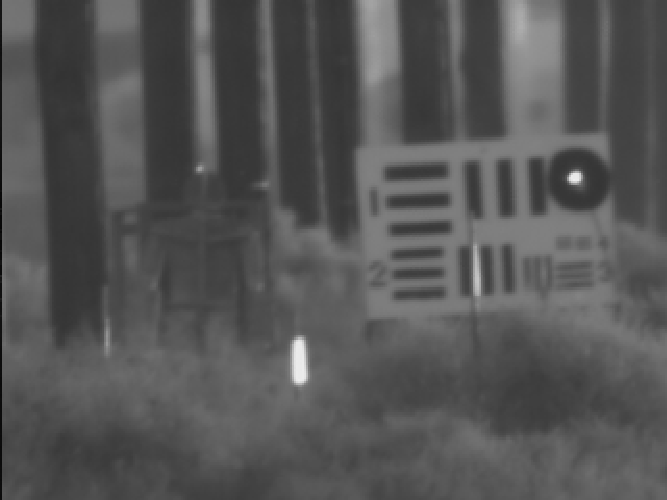}
\end{center}
\caption{Restored images obtained from the different regularizers (30 input frames). From left to right: NLTV, TV, Framelet, Curvelet.}
\label{example6}
\end{figure}

We conduct our experiments on two distinct sequences (acquired during different field trials with different imaging systems), figure~\ref{in1} 
and \ref{in2} show some randomly choosen frames from these sequences. We apply the stabilization algorithm based on the different regularizers presented 
in the previous section to these sequences. We keep the same set of parameters for all tests except the number $N$ of input frames used. We perform the tests 
with $N=30$ and $N=100$ for sequence~1 and with $N=10$ and $N=30$ for sequence~2. Figures~\ref{example3}, \ref{example4}, \ref{example5}, \ref{example6} 
present outputs of the algorithm for each regularizer. Clearly, no big differences can be observed from these results, on these sequences, all regularizers 
perform well. Obviously TV is the fastest version of the algorithm while NLTV is the most efficient one if textures are present in the image. The 
approaches based on frame sparsity can provide a good tradeoff between reconstruction quality and computational speed.

\section{Conclusion}\label{sec:conclusion}
In this paper we investigate the influence of the choice of the regularization term in a recent work dedicated to turbulence stabilization. We tested
four of the most used regularizers in the litterature: total variation, nonlocal total variation, frame sparsity based on framelet or curvelet dictionaries. The results 
obtained on real images show that no major differences can be observed. The ``power'' of the nonlocal total variation appears in the case of textured 
images where it performs better but with the cost of additional computational time.\\

Following these results, we can suggest that, if efficiency is the major aspect which may be reached, a nonlocal total variation regularization must be prefered otherwise 
if the computational speed is critical, the classic total variation is the best option. If both aspects are important, then a sparsity frame based algorithm can 
provides the best tradeoff.
 
\section*{Acknowledgment}
The authors want to thank the members of the NATO SET156 (ex-SET072) Task Group for the opportunity of using the data collected by the group during the 
2005 New Mexico's field trials. This work is supported by the following grants: NSF DMS-0914856, ONR N00014-08-1-119, ONR N00014-09-1-360.

\end{document}